\documentclass{article}
\usepackage{spconf,amsmath,graphicx}

\usepackage{amssymb}
\usepackage{bm,hhline}

\usepackage{booktabs}

\newcommand{\secref}[1]{{Sec.~\ref{#1}}}
\newcommand{\tabref}[1]{Table~\ref{#1}}
\newcommand{\figref}[1]{Fig.~\ref{#1}}

\newcommand{\dist}{{\mathtt{d}}}
\newcommand{\att}{{\mathtt{a}_{\mathcal{S}}}}
\newcommand{\baratt}{{\bar{\mathtt{a}}_{\mathcal{S}}}}
\renewcommand{\Re}{\mathbb{R}}
\newcommand{\xq}{{\bm{x}_q}}
\newcommand{\yq}{{y_q}}
\newcommand{\range}{$\pm$}

\renewcommand{\paragraph}[1]{\vspace{.5mm}\noindent{\bf #1}\hspace{1mm}}

\title{Geometric Mean Improves Loss For Few-Shot Learning}
\name{Tong Wu$^{\dagger\ddagger}$ and Takumi Kobayashi$^{\dagger\ddagger}$}
\address{$\dagger$ University of Tsukuba, Japan\\ $\ddagger$ National Institute of Advanced Industrial Science and Technology, Japan}
\begin{document}
%
\maketitle
\begin{abstract}
    Few-shot learning (FSL) is a challenging task in machine learning, demanding a model to render discriminative classification by using only a few labeled samples. 
    In the literature of FSL, deep models are trained in a manner of metric learning to provide metric in a feature space which is well generalizable to classify samples of novel classes; in the space, even a few amount of labeled training examples can construct an effective classifier.
    In this paper, we propose a novel FSL loss based on \emph{geometric mean} to embed discriminative metric into deep features. 
    In contrast to the other losses such as utilizing arithmetic mean in softmax-based formulation, the proposed method leverages geometric mean to aggregate pair-wise relationships among samples for enhancing discriminative metric across class categories. 
    The proposed loss is not only formulated in a simple form but also is thoroughly analyzed in  theoretical ways to reveal its favorable characteristics which are favorable for learning feature metric in FSL.
    In the experiments on few-shot image classification tasks, the method produces competitive performance in comparison to the other losses.
\end{abstract}
\section{Introduction}
\label{sec:intro}

Few-shot learning (FSL) draws inspiration from the remarkable human ability of robust reasoning and analysis, particularly in scenarios where limited information is available. 
This paradigm has gained significant traction in various applications, e.g., autonomous vehicles and medical analysis, where resource constraints necessitate efficient learning from scarce data.
FSL~\cite{wang2020generalizing} is formulated as a type of machine learning problem where only a limited number of examples with supervised information are available for the target task. Thus, a key challenge in conquering FSL lies in how to efficiently utilize limited data, which has driven research into various approaches to tackle this challenging problem;
FSL methods can be mainly categorized into two types of approaches, \textit{meta learning} and \textit{metric-learning}. 

For rapid adaptation to new tasks with limited data, meta-learning approaches, such as MAML~\cite{finn2017model} and Reptile~\cite{nichol2018reptile}, aim to \emph{learn-to-learn} through a complex two-stage process. They are composed of a meta-training phase for learning the model to adapt quickly toward various tasks and a meta-testing phase to deploy the adaptation ability to new tasks. 
While being potentially flexible, these approaches often suffer from the computation issues that the complex training processes require significant computational resources, involving careful hyperparameter tuning. 

On the other hand, metric learning focuses is applied to construct a feature space where semantically similar samples are closely embedded; in the space equipped with such a favorable metric, even novel samples could be discriminated on the basis of a few number of labeled samples.
The metric-learning approach~\cite{koch2015siamese,bib:PN,bib:NCAloss} 
often produce robust performance across various domains without requiring extensive fine-tuning.
It is also a computationally efficient approach that trains the deep models in a rather straightforward way based on a \emph{loss} function that induces effective metric in deep feature representation.  
Therefore, a loss plays a key role in the metric-based FSL.
Toward better feature metric, losses to train deep models are required to take into account whole training samples, though the FSL losses~\cite{bib:PN,bib:NCAloss} have difficulty in fully paying attention to whole sample distributions.

Thus, in this work, we prose a novel loss to learn effective feature metric via a deep model for FSL.
The proposed method is built upon softmax-based attention weight~\cite{bib:miniIMN,bib:NCAloss} to encode pair-wise relationships among samples.
The proposed method leverages \emph{geometric mean} to efficiently aggregating those pair-wise weights for taking into account broader structure of sample distributions in a deep feature space in contrast to the other FSL losses which pay much attention to rather limited structure and amount of samples, impeding metric learning over whole samples to improve feature discriminativity across class categories.
While it results in a simple loss formulation, our thorough analysis clarifies various characteristics of the proposed loss in theoretical ways which exhibit superior suitability for FSL metric learning in comparison to the other losses.
Our main contributions are summarized as follows.
\begin{itemize}
    \item We propose a novel loss for learning effective metric in FSL by means of geometric mean over softmax-based pair-wise weights which efficiently encode structure of whole samples to facilitate metric learning for FSL. 
    \item We theoretically analyze the proposed loss from various perspectives to reveal favorable characteristics for learning metric of deep features, while revealing interesting connections to the other losses. 
    \item The experimental results on few-shot image classification tasks demonstrate the efficacy of the proposed loss in comparison to the other FSL losses. 
\end{itemize}

\section{Method}
\label{sec:method}

We start with briefly reviewing two representative loss functions for FSL, PN Loss~\cite{bib:PN} and NCA loss~\cite{bib:NCAloss}, then formulate our proposed loss.

\noindent
{\bf Notations.}
Suppose we have a \emph{support} set $\mathcal{S}=\{(\bm{x}_i, y_i)\}_{i=1}^n$ and a target \emph{query} sample $(\xq, \yq)$ for constructing a loss;
an input image $\mathcal{I}$ is embedded into a $D$-dimensional feature vector $\bm{x}\in\Re^D$ via a deep model $\bm{x}=f_\theta(\mathcal{I})$ equipped with trainable parameters $\theta$, and it is also annotated by a class label $y\in\{1,\cdots,C\}$.
We use distance metric denoted by $\dist(\bm{x},\bm{z})$ to measure a discrepancy between two vectors of $\bm{x}$ and $\bm{z}$; it can be specified such as by Euclidean distance $\dist(\bm{x},\bm{z})=\|\bm{x}-\bm{z}\|_2^2$, as will be discussed in \secref{sec:dist}.

\subsection{Metric-based losses for FSL}
\label{sec:fsl_loss}

To embed effective metric into the feature representation $\bm{x}$ for FSL, it is a key process to train the deep model $f_\theta$ on the basis of a loss which works on the distance metric in the feature space.

\subsubsection{PN loss}
\label{sec:PN}
PN loss used in a prototypical network~\cite{bib:PN} is designed so as to push a query $\xq$ toward the cluster of class $\yq$, which naturally leads to the formulation of
\begin{equation}
\ell_{PN}(\xq,\yq|\mathcal{S}) = -\log\frac{\exp(-\dist(\xq,\bm{\mu}_{\yq}))}{\sum_{c=1}^C \exp(-\dist(\xq,\bm{\mu}_{c}))},
\label{eq:PN}
\end{equation}
where a cluster center of the class $c$ is given by $\mu_c = \frac{1}{n_c}\sum_{i|y_i=c} \bm{x}_i$ and the number of samples in the $c$-th class is denoted by $n_c$;
$n=\sum_{c=1}^C n_c$.

In the PN loss \eqref{eq:PN}, each class is represented by the center vector and then a query sample $\xq$ is required to reduce the distance against the class center $\mu_{\yq}$ that it belongs to, in a manner similar to Center loss~\cite{bib:centerloss}.
The loss is formulated by means of softmax comparing the distance to those of the other classes $c\neq y$, so that it induces compact feature representation within a class. 
However, as each class $c$ is described by only a single center $\mu_c$, the PN loss assumes a uni-modal distribution in a class, imposing rather \emph{hard} constraint on feature representation; it is difficult to cope with complicated in-class distribution of multiple modes.

\subsubsection{NCA loss}
\label{sec:NCA}
In contrast to the PN loss, the NCA loss~\cite{bib:NCAloss} is built upon sample-wise relationships more directly.
It is derived from neighborhood component analysis (NCA)~\cite{bib:NCAanal} as
\begin{equation}
\ell_{NCA}(\xq,\yq|\mathcal{S}) = -\log\frac{\sum_{i|y_i=\yq} \exp(-\dist(\xq,\bm{x}_i))}{\sum_{j=1}^n \exp(-\dist(\xq,\bm{x}_j))}.
\label{eq:NCA}
\end{equation}
It aggregates pair-wise relationships (distances) $\dist(\xq,\bm{x}_i)$ in a direct way without resorting to class centers.
In addition, the NCA loss effectively pay much attention to \emph{neighbor} samples due to {\tt log-sum-exp} of
\begin{multline}
    \log\sum_j \exp(-\dist(\xq,\bm{x}_j)) = \\
    -\dist(\xq,\bm{x}_{j^*}) + \log\left[1+\sum_{j\neq j^*} \frac{\exp(-\dist(\xq,\bm{x}_j))}{\exp(-\dist(\xq,\bm{x}_{j^*}))}\right],
    \label{eq:logsumexp}
\end{multline}
where $j^* = \arg\min_j \dist(\xq, \bm{x}_j)$ indicates a nearest-neighbor sample.
Thereby, the (nearest) neighbor samples could dominate the denominator and numerator in \eqref{eq:NCA} since the second term in \eqref{eq:logsumexp} is significantly decayed for far-away samples; in \eqref{eq:NCA}, the numerator pays attention to \emph{in-class} neighbors while the denominator focuses on \emph{global} neighbors on $\mathcal{S}$.
This sample-wise approach can effectively deal with a complicated distribution of even multiple modes in contrast to the PN loss.

On the other way, the emphasis on neighbor samples would make the far-away samples $\{k|\dist(\xq,\bm{x}_k)\gg \dist(\xq,\bm{x}_{j^*})\}$ less contributive to the loss, impeding \emph{whole} samples from enjoying the metric learning; specifically, the metric against those far-away samples is hardly improved in the loss \eqref{eq:NCA}.

\subsection{Proposed loss}
\label{sec:ours}

Toward further effective feature metric, we formulate a metric-based loss by means of \emph{geometric mean}.

We first rewrite the NCA loss \eqref{eq:NCA} into
\begin{equation}
    \ell_{NCA} = -\log \frac{1}{n_{\yq}}\sum_{i|y_i=\yq}\frac{\exp(-\dist(\xq,\bm{x}_i))}{\sum_{j=1}^n \exp(-\dist(\xq,\bm{x}_j))} + \log n_{\yq},
    \label{eq:NCAmean}
\end{equation}
where the second term is just a constant and thus the first term is an intrinsic form of the NCA loss, based on \emph{arithmetic mean} over softmax-based attention weights of
\begin{equation}
\left\{\att(\xq,\bm{x}_i)\triangleq\frac{\exp(-\dist(\xq,\bm{x}_i))}{\sum_{j=1}^n \exp(-\dist(\xq,\bm{x}_j))}\right\}_{i|y_i=\yq}.
\label{eq:attention}
\end{equation}
As discussed above, far-away samples $\{k|\dist(\xq,\bm{x}_k)\gg \dist(\xq,\bm{x}_{j^*})\}$ gain less attention weights, thereby hardly contributing to the arithmetic mean. 

To remedy it, we leverage \emph{geometric mean} to aggregating the softmax-based attention weights by
\begin{multline}
    \ell_{ours}(\xq,\yq|\mathcal{S})  =
    -\log \bigl[{\textstyle\prod_{i|y_i=\yq}}\att(\xq,\bm{x}_i)\bigr]^{\frac{1}{n_{\yq}}}
    \\
    = -\log \left[\prod_{i|y_i=\yq}\frac{\exp(-\dist(\xq,\bm{x}_i))}{\sum_{j=1}^n \exp(-\dist(\xq,\bm{x}_j))}\right]^{\frac{1}{n_{\yq}}}.
    \label{eq:ours}
\end{multline}
While it is a simple modification from \eqref{eq:NCAmean}, the proposed loss \eqref{eq:ours} endows an important characteristic with metric learning that prohibits \emph{any} attention weight $\att(\xq,\bm{x}_i)=\frac{\exp(-\dist(\xq,\bm{x}_i))}{\sum_{j=1}^n \exp(-\dist(\xq,\bm{x}_j))}$ from being close to 0; it is highly contrastive to NCA loss (\secref{sec:NCA}) which could lead to sparse attention weights dominated by neighbor samples. 
Thus, the proposed loss is effective for learning favorable metric to take into account \emph{whole} in-class samples including far-away ones.
It should be noted that the formulation \eqref{eq:ours} is further rewritten into a simpler form of
\begin{equation}
   \ell_{ours} = \frac{1}{n_{\yq}}\sum_{i|y_i=\yq} \dist(\xq,\bm{x}_i) + \log\sum_{j=1}^n \exp(-\dist(\xq,\bm{x}_j)),
   \label{eq:ours2}
\end{equation}
which is computed by using {\tt sum} and {\tt log-sum-exp} functions.

\subsection{Discussion}
\label{sec:discussion}

As described in \secref{sec:ours}, our method can effectively cope with complicated sample distributions by means of sample-wise softmax attention weights while rendering metric learning to \emph{whole} samples; these two points highlight our contrasts to PN loss (\secref{sec:PN}) and NCA loss (\secref{sec:NCA}), respectively.

In addition, we analyze the proposed loss from the following three perspectives, which further clarifies connection not only to PN and NCA losses but also to classification losses.

\subsubsection{Relationship to NCA loss}

The proposed loss \eqref{eq:ours} works as an upper bound of the NCA loss \eqref{eq:NCA} as
\begin{multline}
    \ell_{ours}=-\log \Bigl[\prod_{i|y_i=\yq} \att(\xq,\bm{x}_i)\Bigr]^{\frac{1}{n_{\yq}}} \\
    \geq -\log \frac{1}{n_{\yq}}\sum_{i|y_i=\yq} \att(\xq,\bm{x}_i) =\ell_{NCA},
\end{multline}
which is easily proven by using Cauchy-Schwarz inequality. 
Thus, even in case that the NCA loss is saturated, our loss would be still valid for further learning metrics in a similar way to \cite{bib:Focal}.

To further clarify the mechanism of the proposed loss in comparison to the NCA loss, we analyze them through the lens of loss gradients. 
The gradients of those two losses with respect to $\xq$ are given by 
\begin{align}
\frac{\partial \ell_{ours}}{\partial \xq} &= -\frac{1}{n_{\yq}} \sum_{i| y_i = \yq} \frac{1}{\att(\xq,\bm{x}_i)} \frac{\partial}{\partial \xq}\att(\xq,\bm{x}_i),\label{eq:ours_grad}\\
\frac{\partial \ell_{NCA}}{\partial \xq} &= -\frac{1}{n_{\yq}} \sum_{i|y_i = \yq}\frac{1}{\baratt(\xq,\yq)} \frac{\partial}{\partial \xq}\att(\xq,\bm{x}_i),\label{eq:NCA_grad}
\end{align}
where $\baratt(\xq,\yq)=\frac{1}{n_{\yq}}\sum_{i|y_i=\yq}\att(\xq,\bm{x}_i)$ is an averaged attention weight in the target class $\yq$.
The key difference in the loss gradients (\ref{eq:ours_grad},\,\ref{eq:NCA_grad}) is in the weights for sample-wise gradients $\frac{\partial}{\partial \xq}\att(\xq,\bm{x}_i)$. 
Our loss gradient employs adaptive weights based on the attention $\att(\xq,\bm{x}_i)$; for far-away samples exhibiting less attention, the gradient at the sample is assigned with the higher weight $\frac{1}{\att(\xq,\bm{x}_i)}$, which effectively promotes metric learning in a similar way to mean shift~\cite{bib:MeanShift}.
On the other hand, in the NCA loss, the sample-wise gradients are equipped with a \emph{uniform} weight $\frac{1}{\baratt(\xq,y)}$.
This analysis regarding loss gradients clarifies the efficacy of the proposed loss for learning.

\subsubsection{Relationship to PN loss}

As shown in \eqref{eq:ours2}, our method reduces sum of sample-wise distances $\frac{1}{n_{\yq}}\sum_{i|y_i=\yq}\dist(\xq,\bm{x}_i)$ which, by using $L_2$ distance $\dist=\|\cdot\|_2^2$, is decomposed as
\begin{equation}
\frac{1}{n_{\yq}}\sum_{i|y_i=\yq}\|\xq-\bm{x}_i\|_2^2 = \|\xq-\bm{\mu}_{\yq}\|_2^2 + \frac{1}{n_{\yq}}\sum_{i|y_i=\yq}\|\bm{x}_i-\bm{\mu}_{\yq}\|_2^2.
\label{eq:ours_to_PN}
\end{equation}
While the first term is the distance to the class center $\bm{\mu}_{\yq}$, the target to be minimized in the PN Loss \eqref{eq:PN}, the second term indicates a within-class variance at the class $\yq$.
Therefore, our method minimizes the distance to a class center as in the PN loss, while additionally minimizing the within-class variance to further enhance feature discriminativity; the within-class variance that PN loss lacks is also helpful for learning feature metric from discriminative perspective as in discriminant analysis~\cite{bib:PR}. 

\subsubsection{Relationship to classification loss}
\label{sec:multilabel}

The proposed loss \eqref{eq:ours} is also described by
\begin{equation}
   \ell_{ours} = - \frac{1}{n_{\yq}}\sum_{i|y_i=\yq} \log\frac{\exp(-\dist(\xq,\bm{x}_i))}{\sum_{j=1}^n \exp(-\dist(\xq,\bm{x}_j))},
   \label{eq:ours_multilabel}
\end{equation}
which can be viewed as a \emph{multi-label} softmax loss in a classification framework where a query $\xq$ is categorized into $n$ \emph{pseudo classes} which are respectively represented by$\{\bm{x}_j\}_{j=1}^n$.
The classification is based on $n$ logits of $\{-\dist(\xq,\bm{x}_j)\}_{j=1}^n$ and \emph{multi-hot} labels over $n$ multiple pseudo classes, denoted by 
\begin{equation}
    \hat{\mathtt{p}}_i = \left\{\begin{array}{cc}
    \frac{1}{n_{\yq}} & \text{if } y_i=\yq\\
    0 & \text{otherwise}
    \end{array}\right. \forall i\in\{1,\cdots,n\}, \
    \sum_{i=1}^n \hat{\mathtt{p}}_i = 1.
    \label{eq:multihot}
\end{equation}
Therefore, the loss \eqref{eq:ours_multilabel} is equivalent to cross-entropy between the multi-hot label \eqref{eq:multihot} and the softmax posterior probabilities over the pseudo classes, i.e., the attention weights \eqref{eq:attention}.
From this perspective, our loss enforces the softmax probabilities \eqref{eq:attention} to be close to the multi-hot ones \eqref{eq:multihot}, leading to
\begin{multline}
    \frac{\exp(-\dist(\xq,\bm{x}_i))}{\sum_{j=1}^n \exp(-\dist(\xq,\bm{x}_j))} = 
\frac{\exp(-\dist(\xq,\bm{x}_{i'}))}{\sum_{j=1}^n \exp(-\dist(\xq,\bm{x}_j))}=\frac{1}{n_{\yq}} \\
\Rightarrow \dist(\xq,\bm{x}_i) = \dist(\xq,\bm{x}_{i'}), \forall (i,i')|y_i=y_{i'}=\yq.
\label{eq:medoid}
\end{multline}
This analysis also reveals that our loss pushes $\xq$ toward \emph{medoid} on the distance metric $\dist$.

\subsection{Distance metric}
\label{sec:dist}

We can arbitrarily design basic distance metric $\dist$ used in the loss;
in this literature, Euclidean distance $\dist=\|\cdot\|_2^2$ is commonly utilized to produce favorable performance~\cite{bib:NCAloss,bib:PN}.
In this work, it is formulated based on $L_p$ norm as
\begin{equation}
    \dist_p(\bm{x},\bm{z}) = \sum_{d=1}^D |x_d-z_d|^p.
    \label{eq:dist}
\end{equation}
As discussed in \secref{sec:multilabel}, different types of distance metric, i.e., $p$ in \eqref{eq:dist}, pull a query $\xq$ to different medoids; for $p=2$ (Euclidean distance), $\xq$ is moved toward a simple class mean vector while $p=1$ provides a medoid robust to outliers~\cite{bib:PR}.
We empirically analyze the distance metric in \secref{sec:perf_analysis}.

\begin{table}[t]
    \small
  \centering
  \tabcolsep=1.4mm
  \begin{tabular}{ccccc}
      \toprule
      & \multicolumn{2}{c}{{\textit{mini}ImageNet}} & \multicolumn{2}{c}{{CIFAR-FS}}\\[0.1ex] \cmidrule(lr){2-3} \cmidrule(lr){4-5}
      Loss & 1-shot & 5-shot & 1-shot & 5-shot\\
      \midrule
      BCE~\cite{bib:multilabel_review} & 61.61\range 0.20 & 76.35\range 0.16  & 69.26\range 0.22 & 83.82\range 0.17\\
      ASL~\cite{bib:ASL} & 58.01\range 0.20 & 70.49\range 0.17  & 66.61\range 0.23 & 78.82\range 0.17\\
      \midrule
      Ours & \bf 64.04\range 0.20 & \bf 79.12\range 0.15 & \bf 71.19\range 0.22 & \bf 84.15\range 0.16\\
      \bottomrule
  \end{tabular}
  \caption{Performance comparison (accuracy \%) from the viewpoint of multi-label classification losses (\secref{sec:multilabel}).}
  \label{tab:multilabel}
\end{table}
\begin{figure}[t]
    \small
  \centering
  \begin{tabular}{cc}
    \includegraphics[keepaspectratio,clip,height=0.35\hsize]{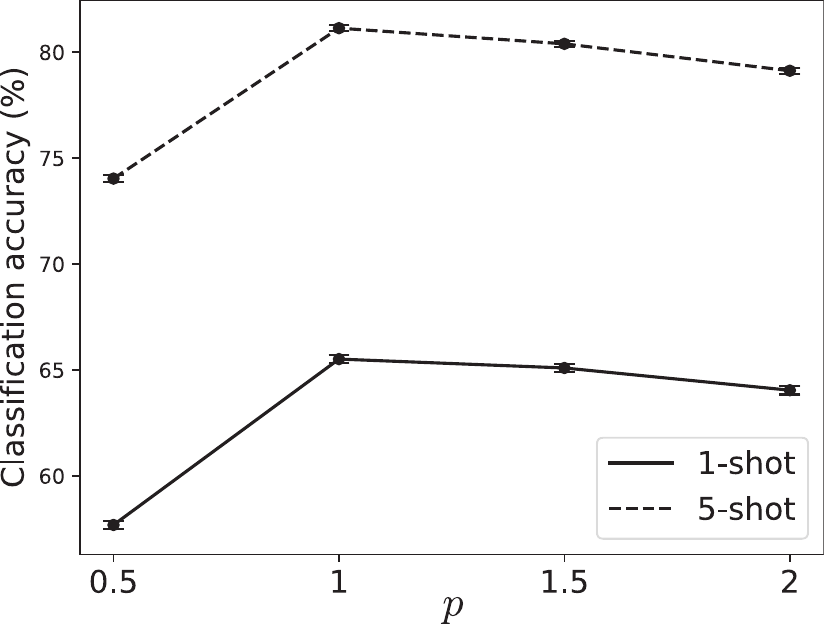}&
    \includegraphics[keepaspectratio,clip,height=0.35\hsize]{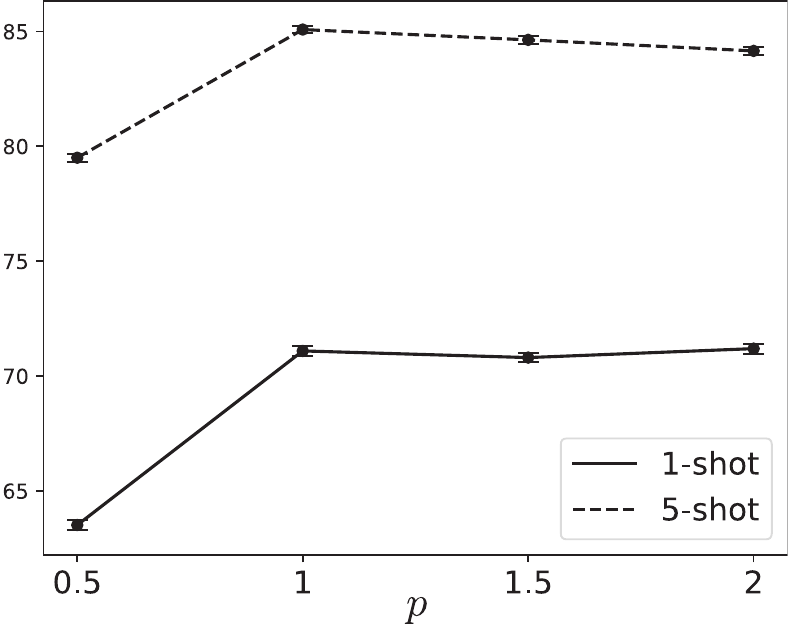}\\[-1mm]
    \textit{mini}ImageNet & CIFAR-FS\\[-3mm]
  \end{tabular}
  \caption{Performance analysis of various distance metric $\dist_p$.} 
  \label{fig:dist}
\end{figure}

\section{Result}
We empirically evaluate and analyze the proposed loss \eqref{eq:ours} on few-shot image classification tasks; we primarily focus on performance in terms of loss functions on an FSL framework.

\begin{table*}[t]
    \small
  \centering
  \begin{tabular}{ccccccc}
      \toprule
      & \multicolumn{2}{c}{{\textit{mini}ImageNet}} & \multicolumn{2}{c}{{CIFAR-FS}} & \multicolumn{2}{c}{{\textit{tiered}ImageNet}}\\[0.1ex] \cmidrule(lr){2-3} \cmidrule(lr){4-5} \cmidrule(lr){6-7}
      Loss & 1-shot & 5-shot & 1-shot & 5-shot & 1-shot & 5-shot\\
      \midrule
      PN~\cite{bib:PN} \eqref{eq:PN}    &
      62.42\range 0.20& 
      79.13\range 0.15&
      67.14\range 0.22&
      82.36\range 0.16&
      66.74\range 0.23&
      82.14\range 0.17\\
      NCA~\cite{bib:NCAloss} \eqref{eq:NCA} & 
      62.68\range 0.20&
      78.93\range 0.54&
      69.20\range 0.21&
      84.24\range 0.16&
      67.21\range 0.22&
      83.77\range 0.16\\
      \midrule
      Ours \eqref{eq:ours}& 
      \bf 65.51\range 0.20 &
      \bf 81.13\range 0.14 &
      \bf 71.09\range 0.22 &
      \bf 85.08\range 0.16 &
      \bf 69.61\range 0.23 &
      \bf 84.04\range 0.16 \\
      \bottomrule
  \end{tabular}
  \caption{Performance results of FSL losses.  
  The best results are highlighted in \textbf{bold}.}
  \label{tab:losscomparison}
\end{table*}

\begin{table*}[t]
    \small
    \centering
\begin{tabular}{ccccccc}
    \toprule
    &\multicolumn{2}{c}{{\textit{mini}ImageNet}} & \multicolumn{2}{c}{{CIFAR-FS}} & \multicolumn{2}{c}{{{tiered}ImageNet}}\\[0.1ex] \cmidrule(lr){2-3} \cmidrule(lr){4-5} \cmidrule(lr){6-7}
    Method & 1-shot &5-shot & 1-shot & 5-shot & 1-shot & 5-shot\\[0.5ex]
    \midrule
    DN4~\cite{li2019revisiting}    & 61.23\range 0.36 & 75.66\range 0.29  & - & - & - & -\\
    CAN~\cite{hou2019cross}        & 63.85\range 0.48 & 79.44\range 0.34  & - & - & \bf 69.89\range 0.51 & \underline{84.23\range 0.37}\\
    Meta-Baseline~\cite{chen2021meta} & 63.17\range 0.23 & 72.96\range 0.17  & \bf 72.00\range 0.70 & 84.20\range 0.50 & 68.62\range 0.27 & 83.74\range 0.18\\
    RFSIC-simple~\cite{tian2020rethinking}& 62.02\range 0.63 & 79.64\range 0.44 & 71.50\range 0.80 & \bf 86.00\range 0.50 & \underline{69.74\range 0.72} & \bf 84.41\range 0.55\\\cmidrule(lr){1-7}
    PN Loss~\cite{bib:PN} \eqref{eq:PN} & 62.42\range 0.20 & 79.13\range 0.15  & 67.14\range 0.22 & 82.36\range 0.16 & 66.74\range 0.23 & 82.14\range 0.17\\
    NCA Loss~\cite{bib:NCAloss} \eqref{eq:NCA}& 62.68\range 0.20 & 78.93\range 0.54  & 69.20\range 0.21 & 84.24\range 0.16  & 67.21\range 0.22 & 83.77\range 0.16\\
    \midrule
    Ours \eqref{eq:ours} & \bf 65.51\range 0.20 & \bf 81.13\range 0.14 & 71.09\range 0.22 & 85.08\range 0.16 & \underline{69.61\range 0.23} & \underline{84.04\range 0.16}\\
    \bottomrule
\end{tabular}
    \caption{Comparison to various FSL methods. 
    For the comparison methods~\cite{li2019revisiting,hou2019cross,chen2021meta,tian2020rethinking}, the reported scores in respective papers are shown.
    Scores falling within the confidence interval of the \textbf{best} are indicated by \underline{underline}.}
    \label{tab:overallcomparison}
\end{table*}

\subsection{Experimental settings}
\textbf{Datasets.} We employ three FSL benchmark datasets. 
The \underline{\textit{mini}ImageNet}~\cite{bib:miniIMN}, derived from the ImageNet, consists of 100 classes with 600 images per class. 
The \underline{CIFAR-FS}~\cite{bib:cifarfs}, a variant of CIFAR-100~\cite{bib:Cifar}, comprises 100 classes with 600 images per class. 
They are split in a way of \cite{ravi2016optimization};  64 classes for training, 16 classes for validation, and 20 classes for test sets, which produce disjoint sets in terms of class categories.
The \underline{\textit{tiered}ImageNet}~\cite{bib:tieredIMN}, based on ImageNet, contains 608 classes which are split into 351 training, 97 validation and 160 test classes.
Input images are resized into $84\times 84$ pixels.

\paragraph{Our FSL framework.}
Following \cite{bib:NCAloss}, we train a deep model $\bm{x}=f_\theta(\mathcal{I})$ by the loss (\secref{sec:method}) computed on mini-batch samples $\mathcal{B}=\{(\bm{x}_i,y_i)\}_{i=1}^n$ randomly drawn from a training set.
Then, the mini-batch set $\mathcal{B}$ is divided into one query sample $(\xq,\yq)\in\mathcal{B}$ and a support subset $\mathcal{S}=\mathcal{B}\setminus(\xq,y)$, and then we compute the loss  repeatedly in a leave-one-out manner over $\mathcal{B}$ as
\begin{equation}
\ell(\mathcal{B}) =
\mathop{ \mathchoice{\lower0.5ex\hbox{\Large E}} {\lower0.2ex\hbox{\large $\prod$}} {\prod} {\prod}}\limits_{(\xq,\yq)\in\mathcal{B}}
\ell(\xq,\yq|\,\mathcal{B}\setminus(\xq,y)).
\end{equation}

\noindent
\textbf{Evaluation protocol.} Then, we evaluate the FSL performance of the trained model $f_\theta$ by following \cite{wang2019simpleshot,rusu2018meta}. 
For simulating few-shot scenarios, we draw $N$-way $K$-shot samples, i.e., $K$ labeled support samples over $N$ novel classes, from a test set which contains no overlapped classes with the training set; 
it is also accompanied by drawing 15 \emph{unlabeled} query samples per class. 
By embedding those $NK$ labeled samples into a feature space via $f_\theta$, a simple classifier is constructed by means of a nearest-mean classification to categorize 15 unlabeled query samples into one of $N$ classes.
We evaluate performance on $N=5$-way $K\in\{1,5\}$-shot scenarios, reporting the averaged classification accuracy with 95\% confidence interval over 10,000 trials.

\paragraph{Model architecture.} We apply a ResNet12~\cite{lee2019meta} 
 as a deep model $f_{\theta}$. 
We train the model from scratch on the training set which is additionally equipped with a linear projection head~\cite{bib:NCAloss} to produce $D=192$-dimensional feature vector $\bm{x}$ for facilitating metric learning only in the training phase; the projection head is detached at the test phase to produce 640-dimensional features used for classification (evaluation).
For training the model, we apply to a mini-batch of 512 samples an SGD optimizer with Nesterov momentum of 0.9, weight decay of 5e-4, and initial learning rate of 0.1 which is decayed by a factor of 10 at 84-th epoch over 120 training epochs including 10 warm-up epochs. 

\subsection{Performance analysis}
\label{sec:perf_analysis}
We analyze the method from various aspects. 

\paragraph{Multi-label classification loss.}
As discussed in \secref{sec:multilabel}, our loss is also viewed from the perspective of a multi-label classification loss.
Thus, we compare our loss \eqref{eq:ours} with the other types of multi-label losses which are widely applied in the classification literature, binary cross-entropy loss (BCE)~\cite{bib:multilabel_review} and asymmetric loss (ASL)~\cite{bib:ASL}; all the methods are equipped with the distance metric $\dist_{p=2}$ in \eqref{eq:dist} for fair comparison.
\tabref{tab:multilabel} reports performance comparison on \textit{mini}ImageNet and CIFAR-FS datasets, demonstrating that the proposed loss outperforms those multi-label classification losses.
The multi-label losses of BCE and ASL mainly focus on sample-wise relationship $\dist(\xq,\bm{x}_i)$ in an individual manner while rather paying less attention to the relationships among whole samples.
In contraast, as discussed in \secref{sec:multilabel}, our loss effectively pushes samples toward in-class medoid, which is favorable for learning effective metric in the FSL framework.

\paragraph{Distance metric $\dist_p$.}
We formulate the distance metric $\dist_p$ in \eqref{eq:dist} based on $L_p$ norms and thus empirically evaluate performance of various $p$ in \figref{fig:dist}.
The results show that $p=1$ produces favorable performance.
The distance metric $\dist_{p=1}$ provides an effective medoid representation in \eqref{eq:medoid} robust against some outlier samples which may be included in the training set.
The robustness contributes to enhancing metric learning.
We apply $\dist_{p=1}$ to the losses in the following experiments.

\paragraph{Comparison to FSL losses.}
In \secref{sec:discussion}, we have analyzed superiority to PN loss \eqref{eq:PN} and NCA loss \eqref{eq:NCA} from theoretical viewpoints.
We qualitatively show performance comparison to those FSL losses in \tabref{tab:losscomparison}; based on the above analysis, all the methods are equipped with the distance metric $\dist_{p=1}$ for fair comparison.
In accordance with the discussion in \secref{sec:discussion}, our loss consistently outperforms PN and NCA losses on various datasets and FSL scenarios.
It should be noted that the proposed loss is as simple as those two methods without increasing computation cost.

\subsection{Performance comparison}

We also show comparison to the other FSL approaches in \tabref{tab:overallcomparison}, though our main focus is the loss function shown in \tabref{tab:losscomparison}.
While our method just works on a loss, it provides competitive performance even in comparison to the other FSL approaches.

\section{Conclusion}

We have proposed a FSL loss based on geometric mean of softmax-based sample-wise attention weights.
While it is formulated in a simple form, our theoretical analysis reveals that the method renders various favorable characteristics to metric learning for FSL in comparison to the other FSL losses.
The experimental results on few-shot image classification tasks empirically demonstrate the efficacy of the proposed loss.

\bibliographystyle{IEEEbib}
\bibliography{template}

\end{document}